\documentclass[]{spie}  
\usepackage{color}
\usepackage[]{graphicx}
\usepackage{amsfonts}

\title{Recovery of underdrawings and ghost-paintings via \\ style transfer by deep convolutional neural networks:  \\ {\LARGE A digital tool for art scholars}}

\author{Anthony Bourached,\supit{a} George H.~Cann,\supit{b} Ryan-Rhys Griffiths,\supit{c} and David G.~Stork\supit{d} 
\skiplinehalf 
\supit{a}Computer Science Department, University College London, London, UK \\
\supit{b}Department of Space and Climate Physics, University College London, London, UK \\
\supit{c}Department of Physics, University of Cambridge, Cambridge, UK \\
\supit{d}Portola Valley, CA 94028 USA}

\authorinfo{Send correspondence to {\tt enquiries@oxia-palus.com}.}

\begin{document} 
\maketitle 

\begin{abstract}
We describe the application of convolutional neural network style transfer to the problem of improved visualization of underdrawings and ghost-paintings in fine art oil paintings.  Such underdrawings and hidden paintings are typically revealed by x-ray or infrared techniques which yield images that are grayscale, and thus devoid of color and full style information.  Past methods for inferring color in underdrawings have been based on physical x-ray fluorescence spectral imaging of pigments in ghost-paintings and are thus expensive, time consuming, and require equipment not available in most conservation studios.  Our algorithmic methods do not need such expensive physical imaging devices.  Our proof-of-concept system, applied to works by Pablo Picasso and Leonardo, reveal colors and designs that respect the natural segmentation in the ghost-painting.  We believe the computed images provide insight into the artist and associated oeuvre not available by other means.  Our results strongly suggest that future applications based on larger corpora of paintings for training will display color schemes and designs that even more closely resemble works of the artist.  For these reasons refinements to our methods should find wide use in art conservation, connoisseurship, and art analysis.
\end{abstract}


\section{INTRODUCTION AND BACKGROUND} \label{sec:Intro}

Many paintings in the Western canon, particularly realist easel paintings from the Renaissance to the present, bear underdrawings and {\em pentimenti} (from Italian, \lq\lq to repent\rq\rq )---preliminary versions of the work created as the artist altered and developed the final design.  The interpretation of such underdrawings is central in addressing numerous problems in the history and interpretation of art, including inferring artists\rq\ praxis as well as interpreting, attributing, and authenticating such works.\cite{Bomford:02,Marijnissen:11}  For example, copies of some paintings have been exposed as forgeries through an analysis of underdrawings revealed through such x-ray imaging;  after all, a forger generally does not have access to the underdrawings and so can duplicate only what is visible to the naked eye.\cite{Lenain:12}  

X-ray and infrared imaging do not directly reveal colors of underdrawings.\cite{Legrandetal:14,Janssensetal:16,deBoer:68}  In most studies, this lack of color information does not pose a major impediment because (it is widely acknowledged) the artist generally uses the same colors throughout the development of the work;  the colors in the primary, visible artwork are likely quite similar to the ones the artist used through its development.  In such cases it is simply the geometric design that is relevant.  For instance, in a computational study of the compositional style of Piet Mondrian, x-ray images of his Neoplastic geometric paintings revealed the designs of prior designs, which the artist ultimately rejected.  Such \lq\lq near miss\rq\rq\ designs, along with the final accepted designs, could be used to train statistical models of Mondrian\rq s compositional principles.\cite{Andrzejewskietal:10}  In that study, x-ray images sufficed because color was of no concern. 

There are, however, some paintings in which the design of the underdrawings are not directly related to that of the visible painting, and these pose a rather different challenge in art analysis.  These are underdrawings in which the entire compositions and designs were painted over by a second unrelated design, which is the visible artwork.  Such hidden artworks or so-called \lq\lq ghost-paintings\rq\rq\ arise when artists re-used canvases, either because they were unsatisfied with their first paintings, or more commonly when they could not afford new canvases.  In some cases the orientation of the hidden design is rotated $90^\circ$ when the artist preferred landscape format to portrait format---or vice versa---for the second painting, as in Picasso\rq s {\em Crouching beggar}.  In some cases the rotation is $180^\circ$ with respect to the visible painting, as for instance Rembrandt\rq s {\em An old man in military costume}.  Prominent paintings that have of such ghost-paintings include:

\begin{itemize}
\item Kazimir Malevich\rq s {\em Black square} (1915)
\item Rembrandt\rq s {\em An old man in military costume} (1630--31)
\item Vincent van Gogh\rq s {\em Patch of grass} (1887)
\item Pablo Picasso\rq s {\em The blue room} (1901), {\em Mother and child by the sea} (1902), {\em The crouching beggar} (1902), {\em Barcelona rooftops} (1903), {\em Old guitarist} (1903--04), and {\em Woman ironing} (1904) 
\item Ren{\'e} Magritte\rq s {\em The portrait} (1935), and {\em The human condition} (1935)
\item Edgar Degas\rq\ {\em Portrait of a woman} (c.~1876--80)
\item Francisco Goya\rq s {\em Portrait of Do{\~n}a Isabel de Porcel} (1805)
\item Leonardo\rq s {\em The Virgin of the rocks} (1495--1508)\footnote{There are two versions of this work:  one in the Louvre and the other in the National Gallery London, which is shown in Fig.~\ref{fig:LeonardoVirgin}, below, and is the focus of our efforts.}
\end{itemize}

\noindent Indeed, as many as $20$ out of $130$ paintings paintings by van~Gogh examined at the van Gogh Museum, Amsterdam, have at least partially completed ghost-paintings.\cite{vanHeugten:95}

There is of course scholarly and general interest in \lq\lq lost\rq\rq\ works, as they give a richer understanding of the artist and his or her oeuvre.\cite{Charney:18}  Art scholars interpret such ghost-paintings for a number of reasons, such as to better understand an artist\rq s career development and choice of subjects (both general and specific), and to learn what an artist did or did not wish to preserve in his oeuvre.\cite{Lowry:15,Faries:03}  It is clear that such tasks are best served by a high-quality digital image of the underdrawing, including features (such as color) not captured by x-radiography and infrared reflectography.  To date, the principal non-destructive method for recovering colors in ghost-paintings is through x-ray fluorescence spectroscopy, which reveals the elemental composition of pigments.  Such measured elemental compositions are then matched to pigment databases so as to infer the likely pigments and their proportions, which in turn indicates the colors in the ghost-painting.\cite{Anithaetal:11,Anithaetal:13,Diketal:08,KammererZoldaSablatnig:03}  This method requires expensive equipment not available in most conservation studios.  Moreover, ultraviolet radiation has shallow penetration power, and may not reveal underdrawings for purely physical reasons.

This, then, is the overriding goal of our work:  to compute a digital image of the ghost-painting as close to the original---in color, form, and style---so the general public and art scholars can better study a hidden artwork, all without the need for such expensive physical sensing equipment.  

In brief, our approach is to transfer the style---specifically the color---from comparable artworks to the grayscale underdrawing.  Related color style transfer in fine art scholarship has been applied to the rejuvenation of fine art tapestries, whose vegetable pigments are fugitive, and thus faded over centuries.  In some cases the cartoons or source paintings survive, where the oil pigments retain the reference colors.  However, tapestry ateliers frequently alter the designs of the works---adding or deleting figures, changing their poses, altering backgrounds, and so forth---and thus a simple overlay of the cartoon or painting design atop the image of a faded tapestry will not lead to a coherent image.  Color transfer with modest tolerance for spatial disparities can produce acceptable \lq\lq rejuvenated\rq\rq\ tapestries, but a full solution will likely require methods based on deep neural networks.\cite{StromJohanssonStork:12} 

We begin in Sect.~\ref{sec:ImageSeparation} with a brief overview of the problem of separating the design of the underdrawing from the x-ray or infrared reflectogram containing the mixtures of the underdrawing and the primary visible image.  We then turn in Sect.~\ref{sec:StyleTransfer} to the problem of mapping of style and colors (learned from representative artworks) to the ghost-painting, an extension and improvement upon earlier efforts,\cite{BourachedCann:19} and enhancing the design of a underdrawing by style transfer from an ensemble of drawings.  We focus on this second stage, where we use deep neural network methods for style transfer.\cite{GatysEckerBethge:16,Gatysetal:17}  We present in Sect~\ref{sec:Results} our image results for two reconstructed ghost-paintings from a single style image and one drawing and from an ensemble of artworks.  We also discuss how such methods can empower art scholars to better interpret these hidden works.  We conclude in Sect.~\ref{sec:Conclusions} with a summary of our results, current limitations and caveats, and future directions for developing this digital tool for the community of art scholars.

\section{IMAGE SEPARATION} \label{sec:ImageSeparation}

An x-ray of a painting containing a ghost-painting reveals the visible and the ghost compositions overlapped.  Thus the first processing stage is to isolate the hidden image from the primary, visible image.  This can be a very challenging computational task, even in simple cases.\cite{MiskinMacKay:00}  The leading algorithmic approach is {\em signal separation} or {\em blind source separation}, which was first developed for separating the sounds of separate sources from a single or multiple audio recordings, as for instance isolating the sound of a single talker from recordings of a noisy cocktail party.\cite{ComonJutten:10,DudaHartStork:01}  The analogous method for images to computationally split the value of each pixel between two candidate images such that each component image is maximally internally correlated and spatially consistent yet maximally uncorrelated and independent from the other image.  The optimization problem works best, at least in theory, when the image-mixing process is simple, for instance that the value of each pixel is the simple sum of the two component images.

Such a simple image mixing does not occur in the x-ray images of fine art paintings, however.  Of course, an x-ray or infrared reflectogram image of a painting bearing a ghost-painting reveals a single image with the two designs overlapped.  However the x-ray opacities of the two component layers of paint do not simply add nor multiply.  Instead, the composition is a nonlinear function that depends upon the particular pigments and binder mixtures (e.g., linseed oil).\cite{Bomford:02}  Moreover, the x-ray attenuation depends upon the thickness of any component layer, even if the reflected appearance of that layer does not.  For instance a layer of chromium yellow pigment of thickness $100\ \mu m$ and one of thickness $300\ \mu m$ will appear nearly identical in reflection, yet affect an x-ray image rather differently.   

Blind source separation is especially difficult with x-ray images of paintings for three reasons:  a)~as mentioned, the image mixing is highly nonlinear, b)~the spatial and chromatic statistics of art vary far more widely than comparable statistics of photographs of natural scenes so spatial coherence is often low, and c)~there are fewer art images available for statistical estimation than numerous easily available corpora of photographs.\index{Dengetal:09}  For these reasons the image separation stage of our work involved hand-editing the underdrawing extracted from x-rays of the target artworks using the {\em GIMP} $2.10$ dodge and burn tool.  Future efforts should lead to progress on this challenging problem in art analysis.

\section{NEURAL STYLE TRANSFER} \label{sec:StyleTransfer}

Style transfer is the task of applying the low-level style (such as color, spatial, and contour statistics) from one or multiple \lq\lq style\rq\rq\ images to the design of another image, the \lq\lq content\rq\rq\ image.  This is an extremely well-studied problem for natural photographs, leading to qualitatively excellent results.\cite{BaeDurand:04,XieTianSeah:07,Zengetal:11}  The metrics for such success of these algorithms are generally qualitative, however:  the modified image should \lq\lq appear as in the style of\rq\rq\ the style images.  For example, one can apply the style of an Impressionist painting to a personal photograph yielding subjectively convincing results.

Style transfer to artworks for scholarly interpretation is a far more difficult problem.  Connoisseurs and art interpreters study the subtlest of details in brushstrokes, paint thickness, contours, colors, and more in their analyses of artworks.  A computed work that merely appears \lq\lq in the style of\rq\rq\ does not suffice for most scholarly analyses beyond coarse, qualitative analyses.  What is needed are systems whose resulting works are nearly indistinguishable in style from genuine works by the artist in question.  Artifacts and spurious features in any computed image must be kept to a minimum, and connoisseurs must learn to identify and account for computational artifacts so as to avoid misinterpreting works.  Our methods, below, are but a first step in this challenging task.

Recently, deep neural networks have been applied to the task of style transfer.\cite{GatysEckerBethge:16,Gatysetal:17}
In broad overview, such networks consist of alternating layers processing \lq\lq content\rq\rq\ image data and \lq\lq style\rq\rq\ data, with cross connections to integrate such information at increasingly higher, more abstract levels.  We present the first use of such networks to recovering ghost-paintings from x-ray images of fine art paintings.

\subsection{TRANSFER OF STYLE FROM A SINGLE ARTWORK} \label{sec:TransferfromOne}

We now turn to the application of neural style transfer to the ghost-painting behind two works by Pablo Picasso:  {\em The old guitarist} (1903--04) and {\em The crouching beggar} (1902).  First, we note that Picasso executed these works early in his career, in his so-called \lq\lq Blue period\rq\rq\ in Barcelona, when he sold very few paintings and sometimes re-used canvases.\cite{Bouvier:19,OBrian:94}  It seems quite likely that his re-use of canvases was based on financial, rather than strictly aesthetic or artistic reasons.  Once Picasso\rq s career became more secure---particularly starting in Paris---he rarely, if ever re-used his canvases.\cite{McCully:97}  Art scholars will gain a deeper, more complete understanding of Picasso\rq s early artistic development once his early ghost-paintings are revealed.

\begin{figure}[h] 
\begin{center}
\begin{tabular}{ccccc}
\includegraphics[width=.17\textwidth]{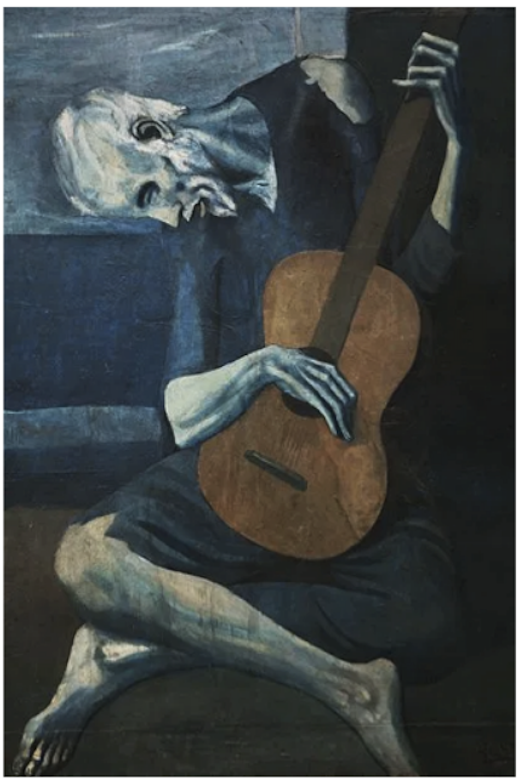} &
\includegraphics[width=.17\textwidth]{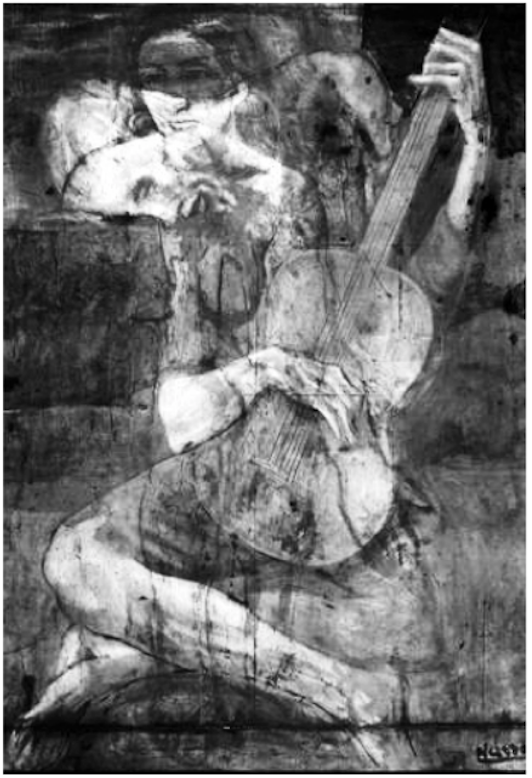} &
\includegraphics[width=.178\textwidth]{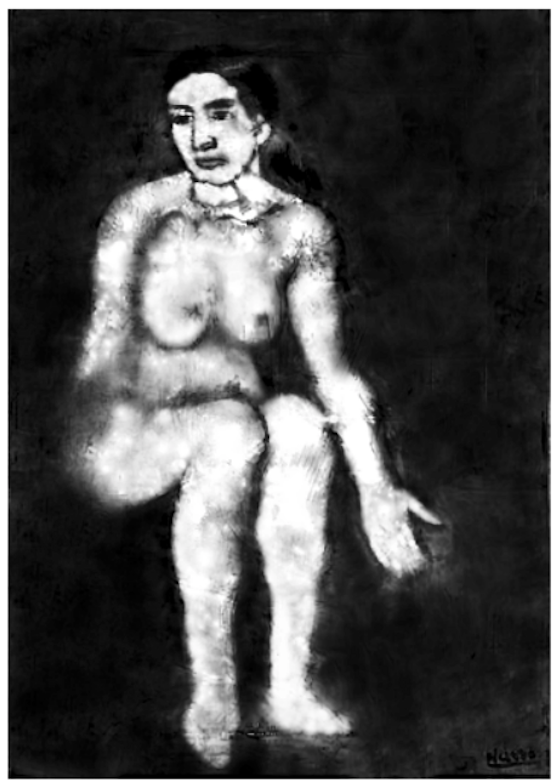} &
\includegraphics[width=.178\textwidth]{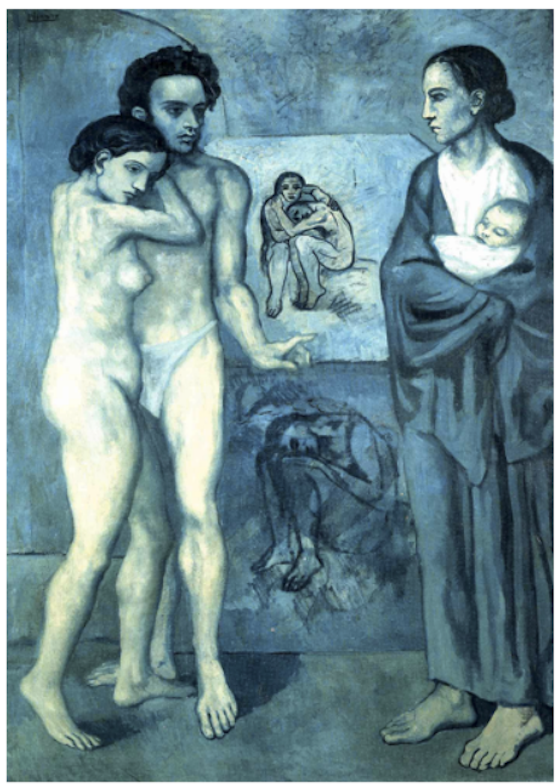} &
\includegraphics[width=.178\textwidth]{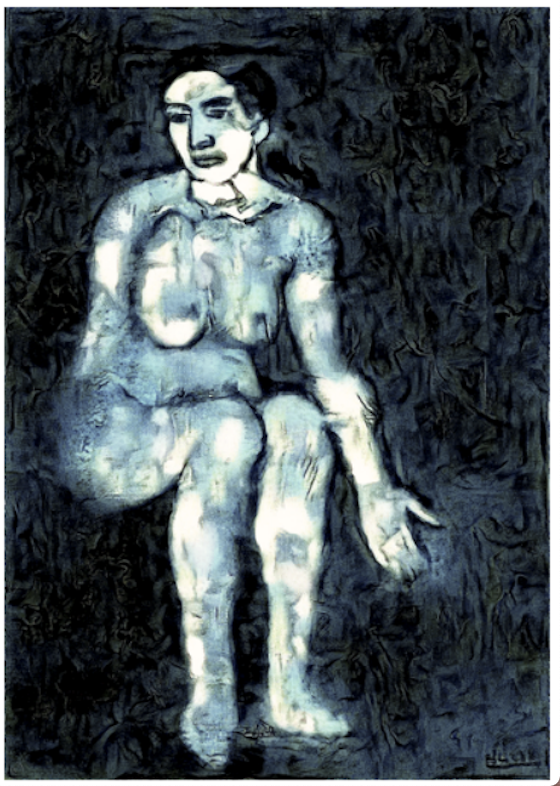} \\
{\small a)} & {\small b)} & {\small c)} & {\small d)} & {\small e)} 
\end{tabular}
\end{center}
\caption{\label{fig:PicassoReconstruction}Works by Picasso used in the computational estimation of the ghost-paintings in {\em Old guitarist}.  a)~{\em Old guitarist} ($122.9 \times 82.6\ cm$), oil on canvas (1903--04), Art Institute of Chicago, b)~x-ray of {\em Old guitarist}, Art Institute of Chicago, c)~the \lq\lq content image,\rq\rq\ i.e., the hand-edited underdrawing of {\em Old guitarist} in grayscale, d)~the \lq\lq style image,\rq\rq\ {\em La vie}  ($196.5 \times 129.2\ cm$), oil on canvas (1903), Cleveland Museum of Art, and e)~the computed portrait, where the color style from d) has been transferred to the design from c) by means of deep neural networks.}
\end{figure}

Figure~\ref{fig:PicassoReconstruction} shows an overview of our approach on the ghost-painting in Pablo Picasso\rq s {\em Old guitarist}.  The x-radiograph reveals the ghost-painting:  a full-length portrait of a seated female nude, much in Picasso\rq s style of that time, with strong outlines and solid, weighty forms.  Of course this x-ray reveals no color information about this hidden work.  Nevertheless, given that it was executed during the artist\rq s blue period (exemplified by {\em Old guitarist} itself), it is natural to transfer style from paintings of figures of the same period, such as {\em La vie} (1903), also executed in Barcelona at that time.  

The recognition network used was the VGG-Network,\cite{simonyan2014very} a Convolutional Neural Network that rivals human performance on common visual object recognition benchmark tasks and was introduced and extensively described. \cite{simonyan2014very} As in \cite{GatysEckerBethge:16} we consider the representations in the feature space of the convolutional layers. We represent the two-dimensional \lq\lq content\rq\rq\ image as a vector, ${\bf x}_C$, and a single \lq\lq style\rq\rq\ images as ${\bf x}_S$, where the pixel dimensions are the same, here $1420 \times 2000$ pixels.  The representations in layer $l$ of the network are $F_l({\bf x}_C)$ and $F_l({\bf x}_S)$, respectively, where $F_l({\bf x}_C) \in \mathbb{R}^{M_l({\bf x}_C) \times N_l}$ and $F_l({\bf x}_S) \in \mathbb{R}^{M_l({\bf x}_S) \times N_l}$.  Here $M_l$ represents layers and $N_l$ are the number of feature maps within a layer $l$, specifically the number of color channels, $N_l = 3$.\cite{GatysEckerBethge:16,Gatysetal:17}

\begin{figure}[h] 
\begin{center}
\begin{tabular}{ccccc}
\includegraphics[width=.115\textwidth,angle=90]{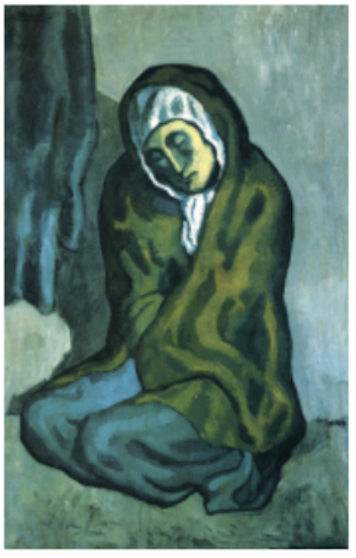} &
\includegraphics[width=.18\textwidth]{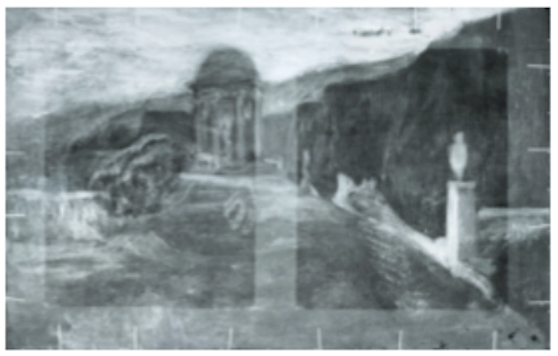} &
\includegraphics[width=.18\textwidth]{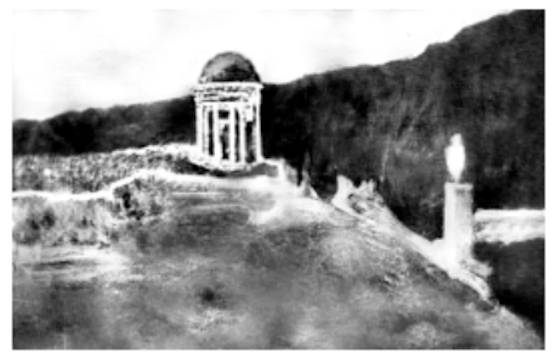} &
\includegraphics[width=.18\textwidth]{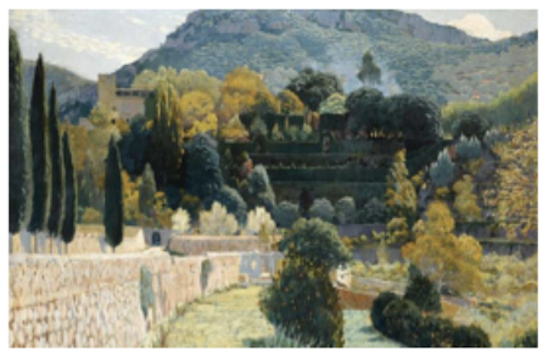} &
\includegraphics[width=.18\textwidth]{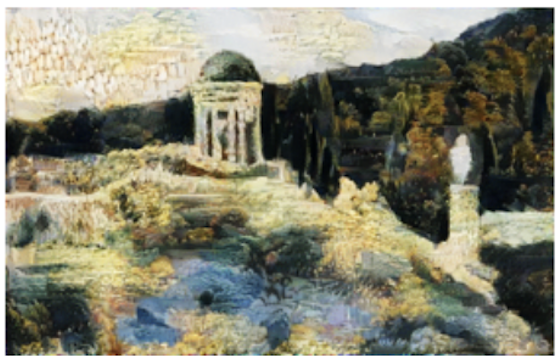} \\
{\small a)} & {\small b)} & {\small c)} & {\small d)} & {\small e)} 
\end{tabular}
\end{center}
\caption{\label{fig:PicassoReconstruction2} a)~Pablo Picasso\rq s {\em The crouching beggar} ($101.2 \times 66\ cm$), oil on canvas (1902), rotated $90^\circ$ counter-clockwise, Art Gallery of Ontario, Toronto, b)~x-ray of {\em The crouching beggar}, Art Gallery of Ontario, Toronto, c)~the \lq\lq content image,\rq\rq\ the hand-edited underdrawing of {\em The crouching beggar}, d)~the \lq\lq style image,\rq\rq\ Santiago Rusi{\~n}ol\rq s {\em Terraced Garden in Mallorca} (dimensions unknown), oil on canvas (1911), and e)~the computed landscape, where the color style from d) has been mapped to the design from c).}
\end{figure}

We denote the stylized final image as $\widehat{\bf x}$, which is obtained by computationally minimizing the following total error or cost function:

\begin{equation} \label{eq:TotalCost}
C_{tot} = \alpha \underbrace{\frac{1}{N_{l_c} M_{l_c}({\bf x}_C)} \sum\limits_{ij} (F_{l_c}(\widehat{\bf x})-F_{l_c}({\bf x}_C))_{ij}^2}_{C_{\rm cont}} + \beta \underbrace{\sum\limits_l w_l E_l}_{C_{\rm style}} ,
\end{equation}

\noindent where $i$ and $j$ are the indices over the component layers and neurons in each component layer, $C_{\rm cont}$ and $C_{\rm style}$ are the content and style cost functions, and $w_l$ and $E_l$ are component errors at each layer.  Training consists of stochastic gradient descent in these error functions.  The relative weights of the two components of the error were set by hand to $\alpha = 10$ and $\beta =40$, which yielded good images. 

\begin{figure}[h] 
\begin{center}
\includegraphics[width=.5\textwidth]{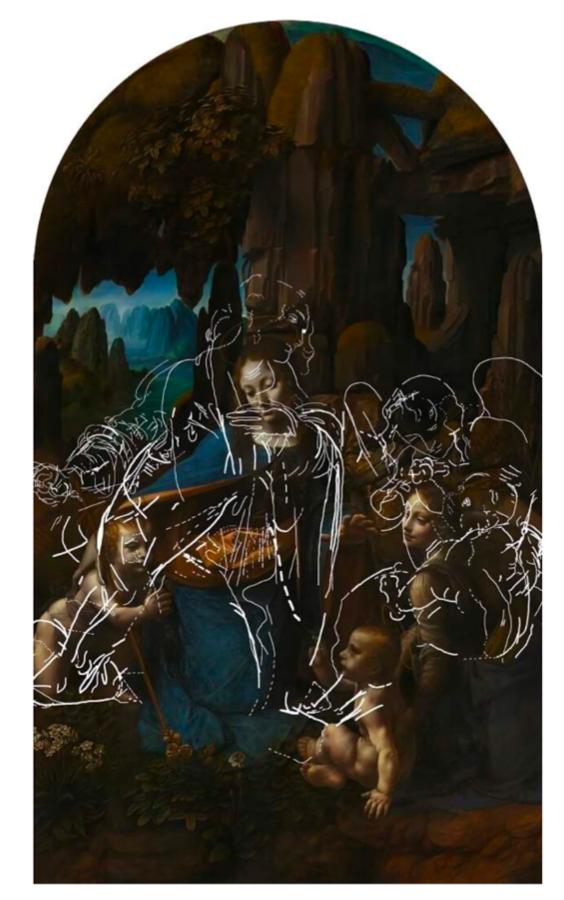} 
\end{center}
\caption{\label{fig:LeonardoVirgin} Leonardo\rq s {\em Virgin of the rocks} ($189.5 \times 120\ cm$), oil on canvas (c.~1499), and negative (white-black reversed) underdrawing, National Gallery London.  We applied style transfer to the underdrawing revealed by x-radiography of this work, shown at the left in Fig.~\ref{fig:LeonardoOverview}.}
\end{figure}

Our compute environment was Google Compute Engine VM Instance, running on {\em NVIDIA} 16GB V100, 8CPU GCP Deep Learning VM.  All software was based on {\em Tensorflow} $1.14$ and {\em CUDA}-$10.0$.  All content and noise images were scaled to $1412 \times 2000$ pixels in $24$-bit RGB.  Hand-edited underdrawing extracted from x-rays performed using the {\em GIMP} 2.10 dodge and burn tool.  As in the Picasso work described above, the content cost coefficient $\alpha = 10$ and style cost coefficient $\beta = 40$, set by hand, worked well.\cite{GatysEckerBethge:16,Gatysetal:17}  Style transfer ran for $10^4$ iterations minimizing the cost in Eq.~\ref{eq:TotalCost}.  We used the default learning rate of $1.0$.  The NST code implemented used a previously trained convolutional neural network from {\em MatConvNet} called the {\em VGG}-19 network---a 19-layer neural network that has been trained on {\tt imagenet-vgg-verydeep-19}, a large dataset of {\em ImageNet} images.\cite{SimonyanZisserman:14}

We shall interpret all our results in Sect.~\ref{sec:Results}.

\subsection{TRANSFER OF STYLE FROM AN ENSEMBLE OF ARTWORKS} \label{sec:TransferfromEnsemble}

The style transfer in each of the two Picasso works above was based on a complete \lq\lq content\rq\rq\ image and a {\em single} \lq\lq style\rq\rq\ artwork.  As such, the method is somewhat limited.  It is extremely rare that a single work is fully representative of the style of an artist or period.  The rare exceptions of this principle, involving color style, include series in which only color is changed, as in Josef Alber\rq s {\em Homage to the square}, Karl Gerstner\rq s {\em Homage to the homage to the square}, Claude Monet\rq s series on {\em Haystacks} and {\em Rouen Cathedral}, and a few others.  For the more general case we must expand our techniques, described in Sect.~\ref{sec:TransferfromOne}, to the application of style from an {\em ensemble} of representative artworks.

Some underdrawings themselves are not complete and do not contain full bounding contours.  In such cases, color style transfer is unlikely to produce acceptable images because the colors will not be sufficiently constrained geometrically.  In order to transfer color from a \lq\lq style\rq\rq\ image or images in such cases, we must first computationally complete boundaries.  This processes is, technically speaking, style transfer but is focused on form and design, rather than color.

 Leonardo completed a mere $36$ or so oil paintings in his career, and thus any underdrawing or ghost-painting is correspondingly very informative to art scholars.\cite{Zollner:17}  Consider the underdrawing revealed through x-radiography of Leonardo\rq s {\em Virigin of the rocks} in the National Gallery London, one of the great masterpieces of Renaissance art, shown in Fig.~\ref{fig:LeonardoVirgin}.  Clearly this underdrawing is preliminary and incomplete, with numerous contours left open. In order to transfer color style to this underdrawing, then, our first step is to compute a more complete design, where the contours are closed and, to the extent possible, preserve the drawing style of Leonardo.
 
 \begin{figure} 
\includegraphics[width=\textwidth]{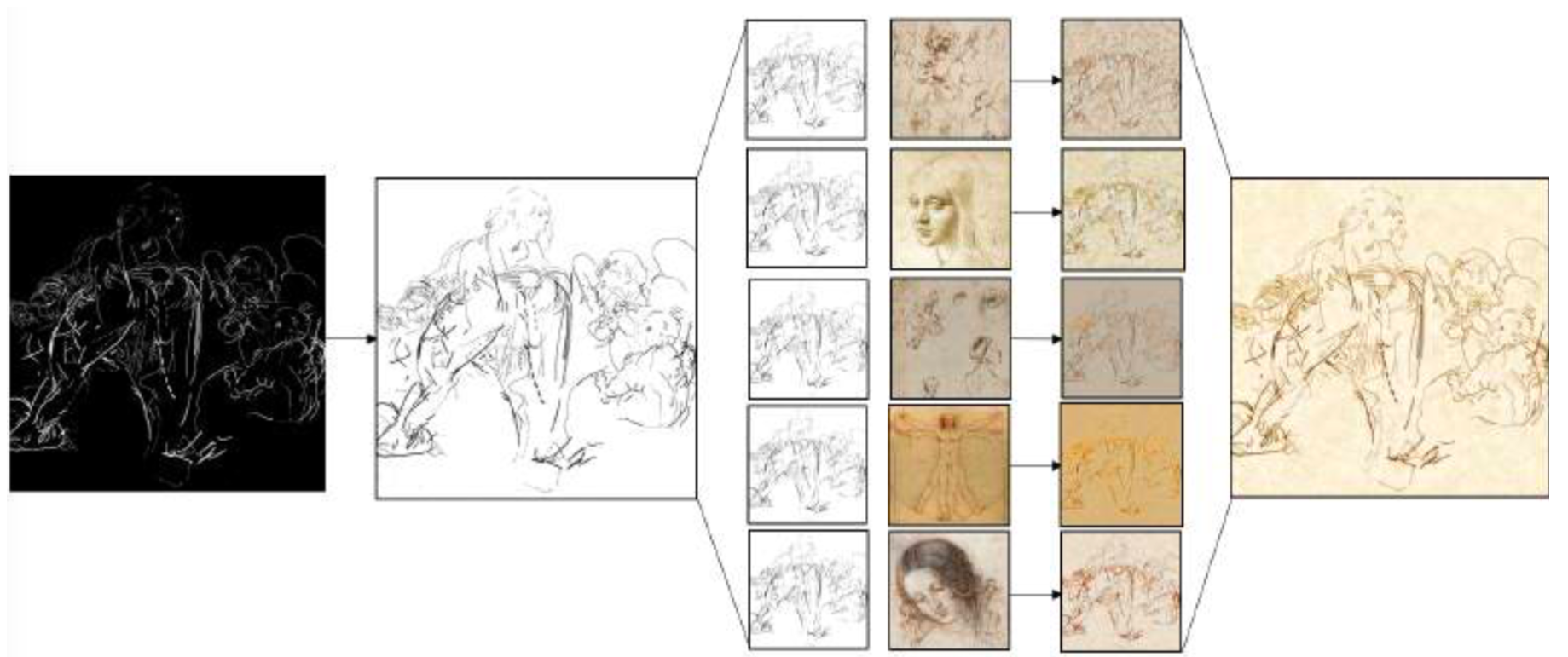}
\caption{\label{fig:LeonardoOverview}An overview of our method for extraction of the ghost-painting in Leonardo\rq s {\em Virgin of the rocks} and style transfer based on an ensemble of Leonardo\rq s works.  At the left is an x-ray of {\em Virgin of the rocks}, and next is a color-inverted version of the x-ray, which facilitates incorporation of Leonardo\rq s traditional drawings and artworks (dark lines on white support).  Next is a set of representative drawings by the artist, each of which leads, through style transfer, to a corresponding contour.  These are then integrated to yield the final, contour-completed model of the underdrawing, at the right.}
\end{figure}

Figure~\ref{fig:LeonardoOverview} shows an overview of our approach to this painting, which relies on deep neural networks.  The x-ray image of the painting is shown at the left, and then a color-inverted (negative) version---dark lines on white support---which facilitates the incorporation of style learned from Leonardo\rq s traditional artworks, in which dark lines are marked on light supports.  We followed the procedure described in Sect.~\ref{sec:TransferfromOne} with a few modifications.  For an ensemble of images, the final recovered image is simply the mean of the individual images:

\begin{equation} \label{eq:AverageImage}
\widehat{\bf x}_f = \frac{1}{m} \sum\limits_{k=1}^m \widehat{\bf x}_k .
\end{equation}

\noindent We shall analyze the resulting work in the next section.

\section{RESULTS AND INTERPRETATIONS} \label{sec:Results}

The right panels in Figs.~\ref{fig:PicassoReconstruction} and \ref{fig:PicassoReconstruction2} show our results on works by Picasso.  The blue flesh, characteristic of Picasso\rq s portraits of this period, is indeed mapped to the recovered work, though a bit less evenly than generally appears in his works of that time.  There are a few passages in which the intensity of the blue conforms to shading, for example the seated figure\rq s shins and wrist.  The recovered landscape behind {\em The crouching beggar} captures the overall color scheme of the reference style image and the colors\rq\ general placement (e.g., sky, trees, fields).  As with the recovered seated portrait, the color application in large passages of relatively uniform color, such as the sky, are somewhat blocking and uneven.  It would be a simple matter to convolve such regions with a broad spatial filter (or use more sophisticated special technique), but a methodologically preferable approach would be to use a large corpora of style images.  Such a principled and data-driven approach reduces significantly the chance that a scholar can bias the method toward images that conform to a prior expectation through choice of \lq\lq style\rq\rq\ artwork.

\section{SUMMARY, CONCLUSIONS, AND FUTURE DIRECTIONS} \label{sec:Conclusions}

We have demonstrated that deep neural networks can be used to transfer style---both color and contour---from representative artworks to an achromatic underdrawing and ghost-painting.  The resulting images of artworks seem, informally at least, to reveal features hard to interpret in achromatic x-ray and infrared reflectogram images.  Clearly this is a proof-of-concept demonstration and should be validated, for instance through ground-truth artworks created in two layers.

We note that the key early step of separating overlapping images from x-ray images of underdrawings automatically is extremely challenging and remains unsolved.\cite{MiskinMacKay:00,DudaHartStork:01}  The general approach that works for linear blind source separation will have to be modified to work in this case because of the physical complexities of x-ray opacity mixing.  In particular two overlapping images do not merely {\em add} opacities to create a final x-ray image.  Presumably methods will build upon blind image separation, but include penalty functions based on the spatial and chromatic statistics of the particular artist in question, and improvements in the nonlinear summation of such component images.  Moreover, the regularization penalties that work best for a given artwork will surely depend upon the artist in question.

Our work advances the recent trend of sophisticated image analysis in the study of fine art.\cite{StoneStork:11}  Such work has solved problems for which traditional connoisseurship was insufficient, such as the claim that Renaissance painters secretly traced optically projected images during the execution of some paintings such as lighting analysis.\cite{Stork:04b,Storketal:11,Johnsonetal:08b} There have been, moreover, promising steps in machine-learning-based neural network-based authentication of artworks, such as characteristic drip paintings by Jackson Pollock, though this work needs to be extended and refined if it is to be used by the art community.\cite{IrfanStork:09}

We conclude that these proof-of-concept results imply that our techniques, suitably extended, refined, and based on larger corpora of style images, will enable art scholars to fill in missing works in an artist\rq s oeuvre and hence gain a more complete view of the artists\rq s development.  Computed works should suffice for studies of large-scale composition, or number and identities of figures.  The grand challenge is to someday recover hidden artworks with such fidelity to the original artistic intention that the computed works are worthy of the careul connoisseurial analysis enjoyed by our \lq\lq visible\rq\rq\ cultural patrimony.\cite{Isolaetal:17,Wangetal:18,Reinhardetal:01,Lietal:17,Selimetal:16,Zhangetal:13,TenenbaumFreeman:00}


\section*{ACKNOWLEDGEMENTS}

We would like to thank the National Gallery London, Art Institute of Chicago, Art Gallery of Ontario, Moretti Fine Art, Ltd., the Italian Cultural Institute in London, the Italian Chamber of Commerce Industry for the UK, Leasys, GRoW$@$Annenberg, and 59~Productions.  The last author would like to thank the Getty Research Center for access to its Research Library, where some of the above research was conducted.

\bibliography{Art}
\bibliographystyle{spiebib}
 
\end{document}